\documentclass[manuscript, screen] {acmart}

\pdfoutput=1

\AtBeginDocument{%
  }

\setcopyright{acmlicensed}
\copyrightyear{2018}
\acmYear{2018}
\acmDOI{XXXXXXX. XXXXXXX}

\acmConference[Conference acronym 'XX] {Make sure to enter the correct
  conference title from your rights confirmation emai}{June 03--05,    
  2018}{Woodstock,  NY}
\acmISBN{978-1-4503-XXXX-X/18/06}

\begin{document}

\title{Parallel Cross Strip Attention Network for Single Image Dehazing}


\author{Lihan Tong\textsuperscript{1} } 
\email{202121301035@jmu.edu.cn}
\affiliation{%
  \institution{Jimei University}
  \city{Xiamen}
  \country{China} 
}
\author{Yun Liu\textsuperscript{2}}
\email{yunliu@swu.edu.cn}
\affiliation{%
  \institution{Southwest University}
  \city{Chongqing}
  \country{China} 
  \institution{, Nanyang Technological University}
  \country{Singapore}
}
\author{Tian Ye\textsuperscript{2}}  
\email{owentianye@hkust-gz.edu.cn}
\affiliation{%
  \institution{The Hong Kong University of Science and Technology (Guangzhou)}
  \country{China} 
}
\author{WeiJia Li \textsuperscript{3} } 
\email{vjiali@jmu.edu.cn}
\affiliation{%
  \institution{Jimei University}
  \city{Xiamen}
  \country{China} 
}
\author{Liyuan Chen \textsuperscript{3} } 
\email{lyuanchen@jmu.edu.cn}
\affiliation{%
  \institution{Jimei University}
  \city{Xiamen}
  \country{China} 
}
\author{Erkang Chen$^{*}$}\thanks{$^{*}$Erkang Chen is the corresponding author.}
\email{ekchen@jmu.edu.cn}
\affiliation{%
  \institution{Jimei University}
  \city{Xiamen}
  \country{China} 
}

\renewcommand{\shortauthors}{Trovato et al. }

\begin{abstract}
The objective of single image dehazing is to restore hazy images and produce clear, high-quality visuals. Traditional convolutional models struggle with long-range dependencies due to their limited receptive field size. While Transformers excel at capturing such dependencies, their quadratic computational complexity in relation to feature map resolution makes them less suitable for pixel-to-pixel dense prediction tasks. Moreover, fixed kernels or tokens in most models do not adapt well to varying blur sizes, resulting in suboptimal dehazing performance. In this study, we introduce a novel dehazing network based on Parallel Stripe Cross Attention (PCSA) with a multi-scale strategy. PCSA efficiently integrates long-range dependencies by simultaneously capturing horizontal and vertical relationships, allowing each pixel to capture contextual cues from an expanded spatial domain.
To handle different sizes and shapes of blurs flexibly, We employs a channel-wise design with varying convolutional kernel sizes and strip lengths in each PCSA to capture context information at different scales.Additionally, we incorporate a softmax-based adaptive weighting mechanism within PCSA to prioritize and leverage more critical features.

We perform extensive testing on synthetic and real-world datasets and our PCSA-Net has established a new performance benchmark. This work not only advances the field of image dehazing but also serves as a reference for developing more diverse and efficient attention mechanisms.
\end{abstract}

\maketitle

\section{Introduction}
Single image dehazing is an image restoration task that aims to estimate the potential haze-free image from the observed hazy image.  It plays a significant role in many fields,  such as outdoor surveillance~\cite{ye2021perceiving}, outdoor scene understanding \cite{sakaridis2018model,sakaridis2018semantic},  and object detection \cite{li2018end,chen2018domain}. 

The haze in an image is often described using a physical scattering model, represented by the formula:

\begin{equation}
    I(x) = J(x) t(x) + A(1 - t(x)),
\end{equation}

Here, $I(x)$ is the hazy image, $J(x)$ is the clear image, $A$ is the global atmospheric light, and $t(x)$ is the transmission map. Traditional methods, such as dark channel prior \cite{dcp}, color attenuation prior \cite{zhu2014single}, and non-local prior \cite{berman2016non}, have proposed various priors to reconstruct high-quality clear images from hazy images. However, relying solely on these priors can lead to reduced robustness in complex real-world scenes \cite{zhang2022deep}.

Advancements in deep learning bring forth numerous CNN-based dehazing methods. Some methods \cite{cai2016dehazenet,ren2016single} explicitly estimate physical parameters, while others learn directly from hazy to clear images through dehazing networks \cite{qu2019enhanced,griddehazenet,qin2020ffa,hong2020distilling,shao2020domain}.

Recently, Transformers find wide application in computer vision \cite{liu2021swin,wang2021pyramid,wang2022uformer,zamir2022restormer}, pushing dehazing performance to new heights. The attention mechanism plays a crucial role in these methods, with a larger receptive field beneficial for capturing long-range relationships \cite{liu2021swin,zamir2022restormer}, and a multi-scale receptive field improving dehazing performance \cite{bai2012image,lei2023multi,shen2023improved}. However, challenges persist:
(I) Convolution-based dehazing models struggle with a limited receptive field, hindering their ability to capture long-range dependencies. Although Transformers excel at this, their computational complexity, proportional to the square of the feature map resolution, makes them unsuitable for pixel-to-pixel dehazing tasks. MB-TaylorFormer \cite{qiu2023mb} suggests using Taylor expansion for linear computational complexity, but the computational cost remains high, requiring additional calculations to correct errors. Some methods confine self-attention area to reduce complexity, but this can lead to suboptimal results \cite{liang2021swinir,wang2022uformer}. Exploring how to capture long-range relationships while keeping computational complexity low remains essential.
(II) Most current models use fixed-size kernels or tokens \cite{qin2020ffa,liu2021swin,li2017aod,zamir2022restormer}. Further optimization is needed to enable models to effectively adapt the receptive field size for various haze patterns.

To address the first challenge, we introduce PCSA. PCSA simultaneously captures vertical strip features and horizontal strip features, efficiently fusing them to achieve lightweight long-range feature extraction. Specifically, by integrating horizontal and vertical cues, each pixel gains awareness of its surroundings based on its central position, effectively capturing long-range dependencies within the image. This approach reduces computational complexity and maintains model lightweightness by conducting feature extraction along two directions for each pixel.

To tackle the second challenge, we design the PCSA model with a multi-scale  strategy (PCSAM). We divide the input feature map into multiple groups by channels, applying PCSA with varying stripe sizes and convolutional kernel dimensions in each group. We then merge the features from horizontal and vertical stripes along the channel dimension. This strategy enhances features at different scales and adapts perfectly to various blur sizes.

PCSA offers several key advantages: (a)It performs horizontal and vertical band attention in parallel, quickly aggregating information from both directions while maintaining low computational complexity.
(b) The channel-wise multi-scale  strategy dynamically gathers information through multi-scale receptive fields, adapting to blurs of various sizes and improving performance. (c) It efficiently fuses horizontal and vertical band features along the channel dimension, enhancing features with lightweight parameters.

In summary, our main contributions are as follows:
\begin{itemize}
 
\item We introduce the PCSA module, which performs horizontal and vertical band attention in parallel and fuses features along the channel dimension, balancing the capture of long-range relationships with reduced computational complexity.

\item The PCSA module uses multi-scale horizontal and vertical band attention to dynamically capture contextual information, perfectly adapting to blurs of different sizes.

\item Comprehensive experimental results on both synthetic and real dehazing datasets demonstrate that PCSA-Net achieves state-of-the-art (SOTA) performance.
 
\end{itemize}
\section{Related Work}
\subsection{Image Dehazing}
Single image dehazing can be divided into two main approaches: prior-based methods and learning-based methods.  Prior-based methods include techniques like contrast maximization \cite{tan2008visibility},  dark channel prior \cite{dcp},  and color attenuation prior \cite{zhu2014single}.  These methods tend to have limited robustness because they rely on specific scene characteristics.  With the advancement of Convolutional Neural Networks (CNNs),  a variety of CNN-based methods for image dehazing have been proposed \cite{cai2016dehazenet,    zhang2018densely,    li2017aod}.  FFA-Net \cite{qin2020ffa} introduced a Feature Attention (FA) block that combines channel attention with pixel attention,  achieving impressive Peak Signal-to-Noise Ratio (PSNR) and Structural Similarity (SSIM) by fusing features at different levels.  AECRNet \cite{wu2021contrastive} presented a novel loss function based on contrastive learning,  using information from hazy and clear images as negative and positive samples,  respectively,  to achieve higher metrics.  However,  CNN-based models often struggle to learn long-range dependencies.  Transformers effectively capture long-range dependencies but face the issue of quadratic complexity relative to spatial resolution \cite{liu2021swin,    liang2021swinir,    wang2022uformer,    zamir2022restormer}. 
\subsection{Efficient Self-attention}
Attention mechanisms play a crucial role in the field of computer vision.  A multitude of attention modules have been proposed for image dehazing,  achieving surprisingly impressive results \cite{griddehazenet,    chen2023msp}.  PCFAN \cite{zhang2020pyramid} utilizes the complementarity between features of different levels through a pyramid approach combined with a channel attention mechanism for image dehazing.  GridDehazeNet \cite{griddehazenet} implements an attention-based multi-scale estimation,  effectively alleviating the bottleneck issues often encountered in traditional multi-scale methods.  FFA-Net \cite{qin2020ffa} combines channel attention and pixel attention mechanisms to treat different features and pixels unequally,  providing additional flexibility for handling various types of information.  MSAFF-Net \cite{lin2022msaff} employs a channel attention module and a multi-scale spatial attention module to consider areas with fog-related features.  These attention modules have all enhanced the performance of image dehazing models. 
Recently,  Transformers have been widely applied to various computer vision tasks and have subsequently made significant contributions to the progress of advanced visual tasks \cite{liu2021swin,    wang2021pyramid,    touvron2021training,    carion2020end}.  However,  the computational complexity of Transformers is quadratic with respect to the resolution of feature maps,  making them unsuitable for pixel-to-pixel tasks like dehazing.  Swin Transformer \cite{liu2021swin} introduces a hierarchical Transformer,  which brings higher efficiency by limiting self-attention computations to non-overlapping local windows.  Restormer \cite{zamir2022restormer} proposes an efficient Transformer model that captures long-range pixel interactions by making key designs in multi-head attention.  Stripformer \cite{tsai2022stripformer} reweights image features horizontally and vertically to capture different directional blurring patterns,  using less memory and computational cost. However, these methods still face the challenge of quadratic complexity in regions or strips. Compared to the aforementioned methods, the proposed parallel strip attention module efficiently integrates information both horizontally and vertically simultaneously. This module, combined with a feature fusion module in the channel dimension, offers advantages such as high efficiency, a large receptive field, and dynamic feature fusion.

\section{Method}

\begin{figure*}[t!] 
    \centering
   
    \includegraphics[width=0.6\linewidth] {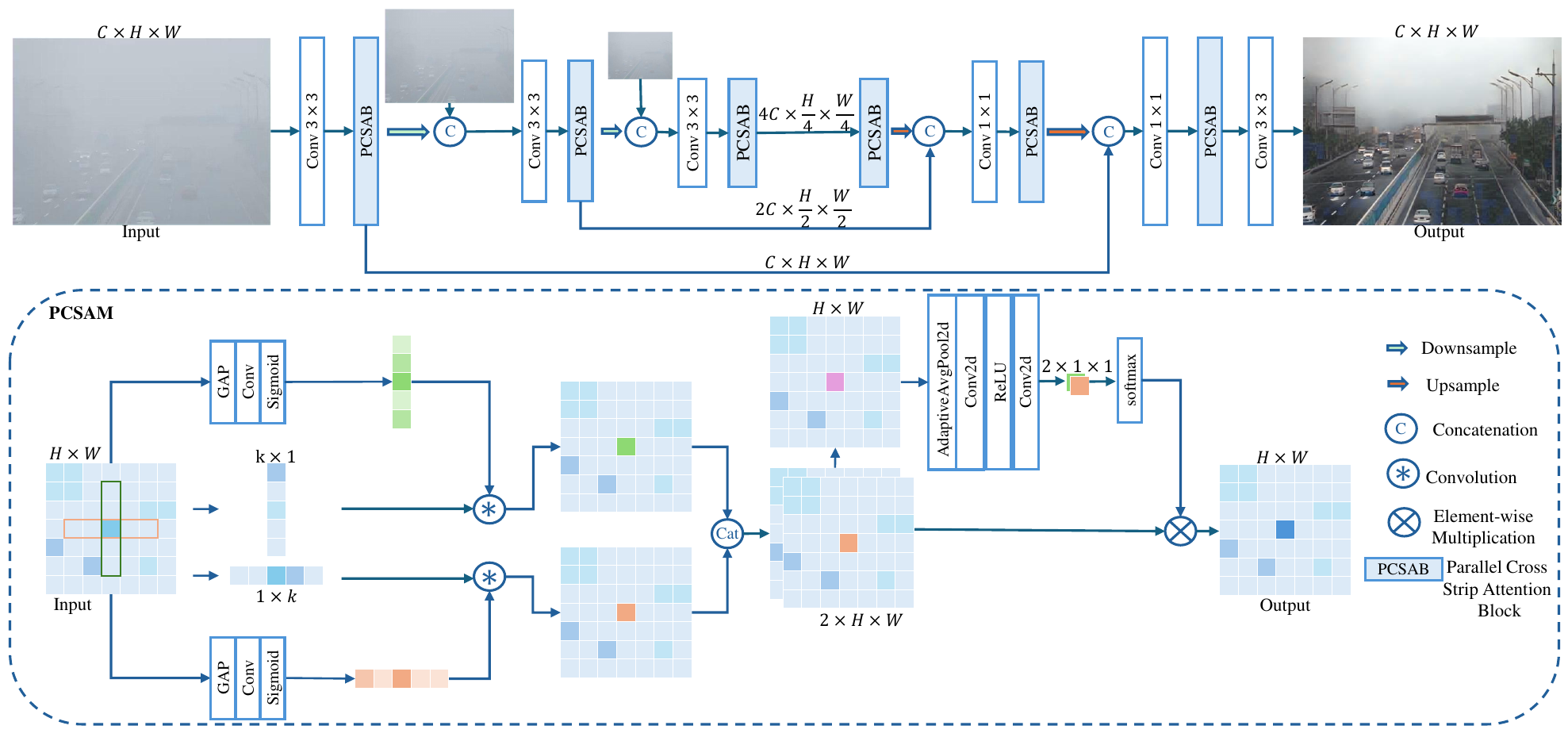}
    \caption{ The Overview of our Parallel Cross Strip Attention  Network architecture.  We give details of the structure and configurations in
Section III. }
\vspace{-0.4cm}
 \label{network}
\end{figure*}

\begin{figure*}[t!] 
    \centering
\includegraphics[width=0.6\linewidth] {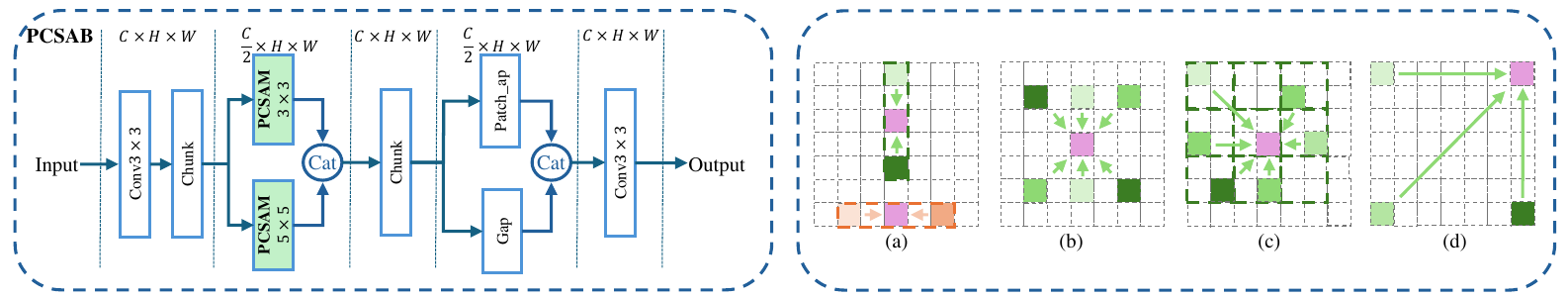}
    \caption{The box on the left shows the structure of the Parallel Cross Strip Attention Block (PCSAB),  which uses a parallel structure for higher execution efficiency.  The PCSAM is the core module of PCSAB,  which is explained in detail in the methods section.  The box on the right summarizes the basic principles of various attention mechanisms.  (a)  our PCSA,  (b) Strip attention \cite{tsai2022stripformer},  (c)  bi-level routing attention \cite{zhu2023biformer},  (d) self-attention.  By comparison,  it can be concluded that our PCSA is the most efficient. }
\vspace{-0.4cm}
 \label{PCSAB}
\end{figure*}

\subsection{Overall Architecture}

To make the model both efficient and capable of capturing long-range relationships,  we propose an encoder-decoder structure that uses PCSAB as its basic building block.  As shown in the Fig.  \ref{network},  we employ three encoders and three decoders,    deploying PCSAB at each scale,  with PCSAM serving as the core module of PCSAB.  To create a compact dehazing model and enhance its efficiency,  we apply a 4x downsampling operation.  To prevent the loss of important information,  we use  PCSAB for feature extraction before each 4x downsampling step.  To reduce training difficulty \cite{mao2021deep,    tu2022maxim} and strengthen the exchange of information between layers \cite{cui2022selective},  we employ skip connections at both the feature and image levels to assist with training.  The entire model achieves surprising metrics and visual effects with just six PCSABs. 
\subsection{Parallel Cross Strip Attention Module}
Our objective is to design a lightweight model that efficiently and flexibly captures relationships across various scales.  Hazy images captured in natural scenes often exhibit irregular and non-uniform haze.  The PCSA   is capable of capturing multi-scale strip features in both horizontal and vertical directions.  It employs intra-strip markers to detect haze of different directions and magnitudes and reweights the features within the strips to adapt seamlessly to fog of diverse sizes and intensities.  During the feature fusion process,  we enhance crucial features on the channel dimension,  highlighting those with determinative roles,  thus ensuring an efficient and comprehensive integration of bidirectional strip features.

\subsubsection{Parallel Cross Strip Attention}
For a given hazy image $  I \in R ^{H \times W \times C} $,    where H,    W and C represent the height,  width,  and number of channels respectively,  we first extract shallow features in both horizontal and vertical directions using global average pooling,  a $ 1 \times 1 $ convolutional layer,  and the Sigmoid function.  This process generates lightweight attention weights.  The formula is as follows:
\begin{equation}
\begin{split}
A^{\perp} = \sigma(Conv_{1 \times 1}(GAP(I))),    \\
A^{\parallel} = \sigma(Conv_{ 1 \times 1 }(GAP(I))),    
\end{split}
\end{equation}
In the formula,  $ \sigma $ represents the Sigmoid function,  $Conv_{1 \times 1}$ represents a $1 \times 1$  convolutional layer  and GAP stands for global average pooling.   Here $A^{\parallel} \in R^{K \times 1}$,    where  K refers to the length of the strip.  Unlike the Stripformer \cite{tsai2022stripformer},  which divides the strips based on the image width $W$ our strips have a length $K$ that's  smaller than $W$, and the size of $K$ can be flexibly controlled,  which will be detailed in a subsection. Then we parallel obtain the strip features in the horizontal direction and the vertical direction. Here's the formula:
\begin{equation}
    \begin{split}
        I^{\perp}_{h,    w,    c}=\sum\limits_{k=0}^{K-1}A^{{\perp}}_k I_{h-{\lfloor \frac{K}{2} \rfloor}+k,    w,    c} \\
        I^{\parallel}_{h,    w,    c}=\sum\limits_{k=0}^{K-1}A^{\parallel}_k I_{h,    w- \lfloor \frac{K}{2} \rfloor+k,    c}
    \end{split}
\end{equation}
Inspired by previous work \cite{liu2021swin,    zamir2022restormer},  expanding the receptive field is beneficial for image restoration.  To broaden the receptive field while maintaining the model's efficiency,  we do not generate a self-attention weight map that matches the size of the input image and weigh it through matrix multiplication.  Instead,  we focus on each pixel and aggregate information through convolution.  The feature maps obtained from horizontal strip attention (HSA) and vertical strip attention (VSA) can be represented as follows:
 \begin{equation}
 \begin{split}
    \hat{I}^{\perp}=VSA(I),    \\
     \hat{I}^{\parallel}=HSA(I),    
 \end{split}
 \end{equation}
 Where HSA stands for Horizontal Strip Attention,  and VSA represents Vertical Strip Attention. To efficiently combine features from both horizontal and vertical directions,  we concatenate and sum the feature maps along the channel dimension from both directions.  Then,  we further integrate the features through adaptive average pooling,  $1 \times 1$ convolution,  ReLU activation function,  and Softmax function.  Finally,  we multiply the integrated feature map with the bidirectional strip input and sum the results to get the final output.  The formula is as follows:
 \begin{equation}
 \begin{split}
      \hat{I}=sum(cat(VSA(I),    HSA(I))),    \\
     M = Conv_{1 \times 1}(ReLU( Conv_{1 \times 1}(Adap(\hat{I}))),    \\
     Output =sum( \hat{I} \times M),    \\
     Output =Fusion(VSA(I),    HSA(I)),    
 \end{split}
  \end{equation}
 Where sum refers to the operation of summing along the channel dimension,  and Fusion denotes the module we designed for combining horizontal and vertical strip features. 
 \subsubsection{Multi-scale  Strategy}
 Inspired by previous work \cite{bai2012image,    lei2023multi,    shen2023improved},  having a multi-scale receptive field helps with image restoration.  To capture multi-scale receptive fields,   we propose a Multi-scale  Strategy. Specifically,  for the input image $I \in R^{H \times W \times C}$
We evenly divide it into two parts along the channel dimension.  Then,  we perform horizontal and vertical strip attention in parallel with different $K$ values. Additionally,  we set convolutional kernels of varying sizes in different PCSAs to flexibly capture features of various sizes. The formula is as follows:
\begin{equation}
\begin{split}
       PCSA_1(I) = Fusion(VSA^{k1}_{Conv 3\times3}(I),    HSA^{k2}_{Conv 3 \times 3}(I)),    \\
    PCSA_2(I) = Fusion(VSA^{k1}_{Conv 5\times5}(I),    HSA^{k2}_{Conv 5 \times 5}(I)) ,  \\
    PCSAM = Cat(PCSA_1(I) ,    PCSA_2(I) ) ,    
\end{split}
\end{equation}

\subsection{Loss Function}
To make full use of negative samples,  we adopt Contrastive Regularization (CR) \cite{wu2021contrastive},  which is based on contrastive learning.  We use the hazy image as the negative sample and the clear image as the positive sample,  with the dehazed image serving as the anchor: $L_{CR}=CR( \text{PCSA-Net(I)},    J_{gt},    I)$ Ultimately,  we combine the loss function derived from Contrastive Regularization with the L1 loss function to create the final loss function. 
\begin{equation}
    L_{total}=L_1(\text{PCSA-Net}(I),  J_{gt}) + \lambda L_{CR} .
\end{equation}
Our experiments shows that the best results are achieved when $\lambda_1=0.2$.

\begin{figure*}[t!] 

    \centering
    \includegraphics[width=0.8\linewidth] {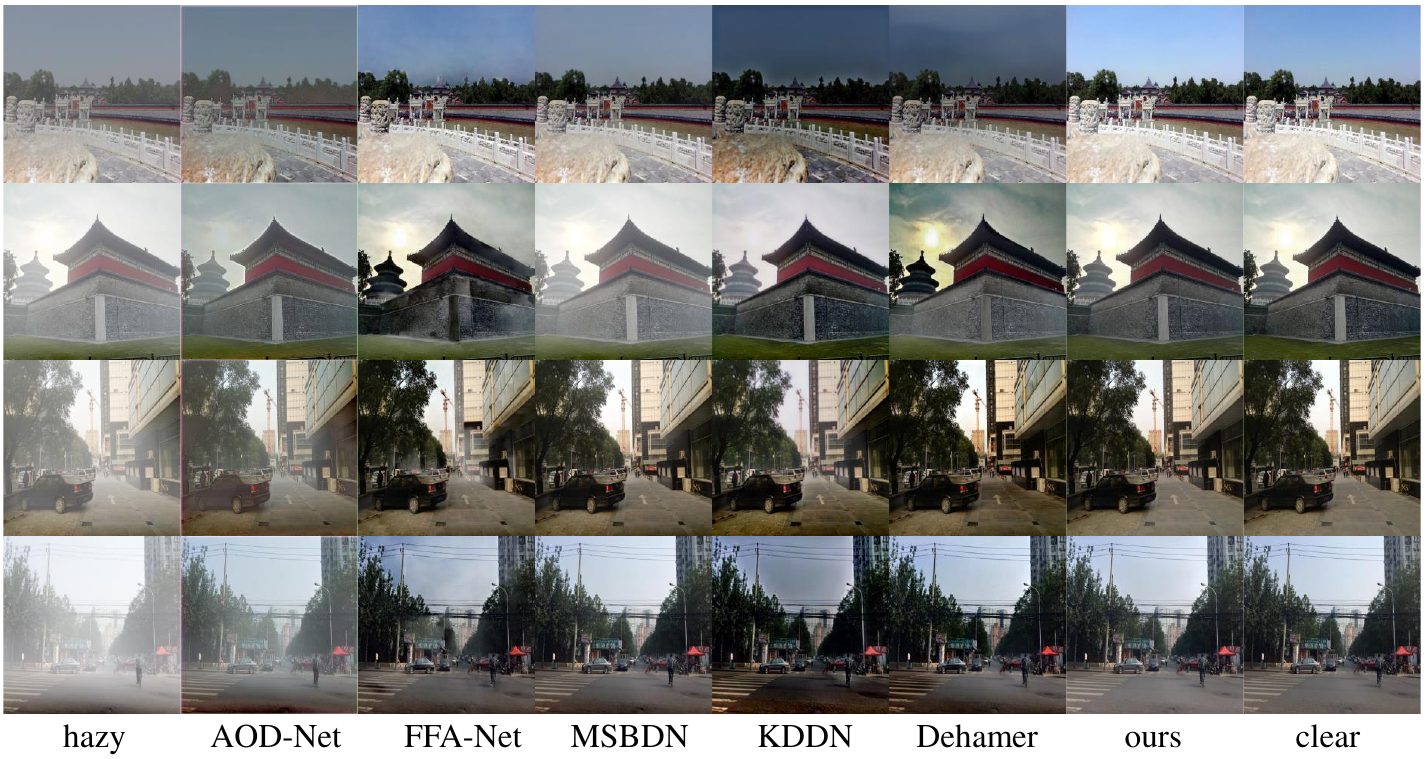}
    \caption{ Visual results comparisons on Haze4K \cite{li2019benchmarking} dataset.   Zoom in for best view. }
 \label{Fig:Haze4K}
\end{figure*}

\begin{figure*}[t!] 
    \centering
    \includegraphics[width=0.8\linewidth] {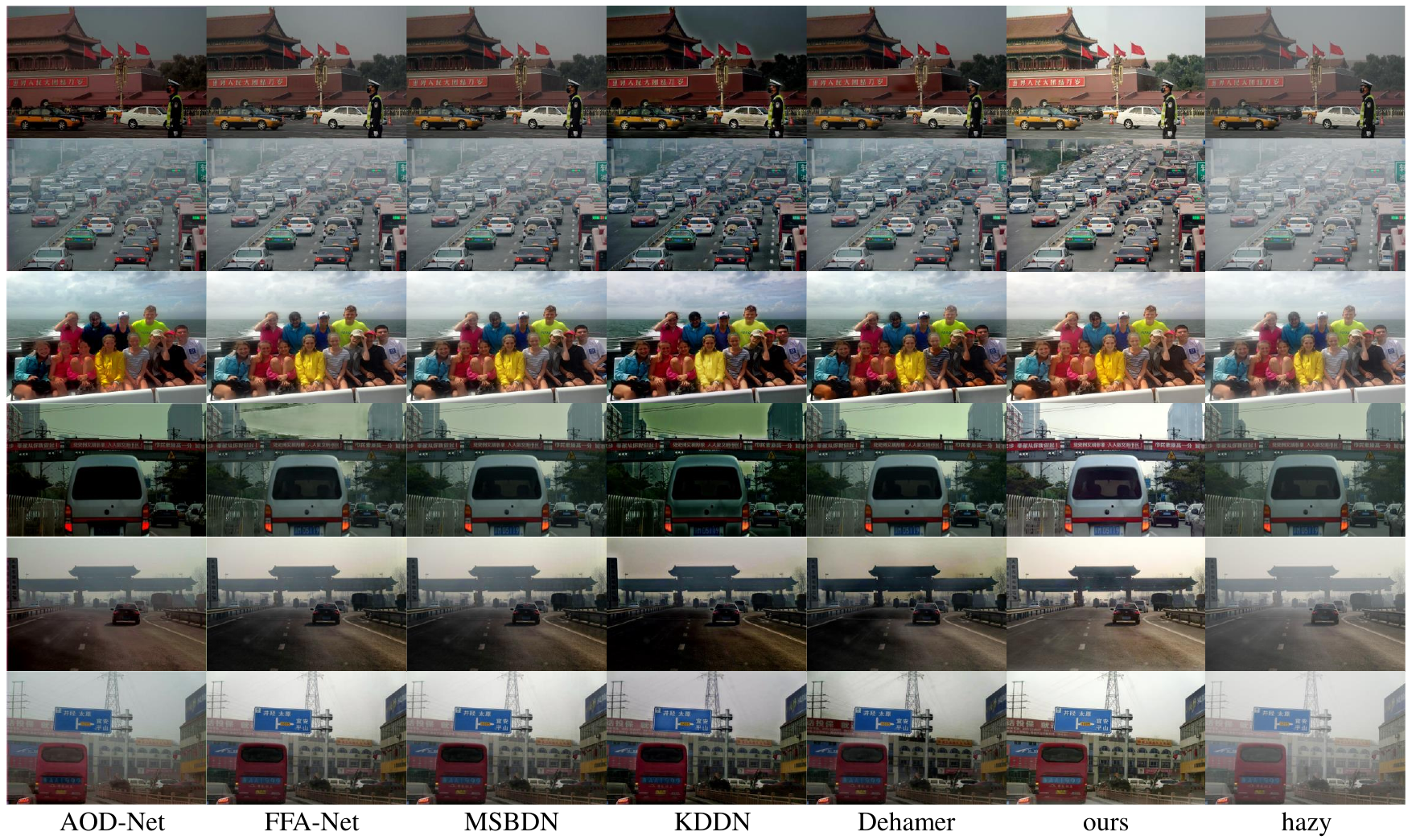}
    \caption{ Visual results comparisons on real-world hazy images.  The samples from the RTTS dataset~ \cite{li2019benchmarking}.  Zoom in for best view. }
 \label{Fig:RTTS}
\end{figure*}

\begin{table}[t] 
	\caption{Quantitative comparisons with SOTA methods on the RESIDE-Indoor~ \cite{li2019benchmarking} and Haze4K \cite{liu2021synthetic} datasets. }
	\label{performance_SOTS}
	\centering
 \resizebox{0.4\textwidth}{!}{
	\begin{tabular}{c|cc|cc|c}
		\hline
		Method & \multicolumn{2}{c|}{RESIDE-IN \cite{li2019benchmarking}} & \multicolumn{2}{c|}{Haze4k \cite{liu2021synthetic}} & \# Param \\ 
		\cline{2-5}
		 & PSNR (dB) & SSIM & PSNR (dB) & SSIM &  \\ 
		\hline
		(ICCV'17) AOD-Net \cite{li2017aod} & 19.82 & 0.818 & 17.15 & 0.83 & 2.0K \\
		(ICCV'19) GridDehazeNet \cite{griddehazenet} & 32.16 & 0.984 & - & - & 1.0M \\
		(AAAI'20) FFA-Net \cite{qin2020ffa} & 36.39 & 0.989 & 26.96 & 0.95 & 4.7M \\
		(CVPR'20) MSBDN \cite{dong2020multi} & 33.79 & 0.984 & 22.99 & 0.85 & 31.4M \\
		(CVPR'20) KDDN \cite{hong2020distilling} & 34.72 & 0.985 & - & - & 6.0M \\ 
		(CVPR'21) AECR-Net \cite{wu2021contrastive} & 37.17 & 0.990 & - & - & 2.6M \\ 
		(CVPR'22) Dehamer \cite{guo2022image} & 36.63 & 0.988 & - & - & - \\
		(ECCV'22) PMNet \cite{ye2021perceiving} & 38.41 & 0.990 & 33.49 & \textbf{0.98} & 18.9M \\
		\hline
		PCSA-Net(Ours) & \textbf{39.40} & \textbf{0.991} & \textbf{33.76} & \textbf{0.98} & 20.4M \\
		\hline
	\end{tabular}
 }
	\vspace{-0.3cm}
\end{table}

\begin{table}[t] 
	\caption{Ablation study of our PCSA-Net on the Haze4k Dataset~ \cite{liu2021synthetic}. }
	\label{ablation_netwrok}
	\footnotesize
	\centering
\resizebox{0.3\textwidth}{!}{\begin{tabular}{cc|cc}
		\toprule[1pt] 
		Model & PSNR (dB) & SSIM \\
		\hline 
            Base(U-Net)   &25.76 &0.92\\
		Base+HSA   &31.76 &0.96\\
		Base+VSA  &31.94 &0.96\\
            Base+HSA+VSA &32.96 &0.97  \\
            Base+PCSA  &33.27 &0.97\\
		Base+PCSAM (Full)  &\textbf{33. 76} &\textbf{0.98} \\

		\bottomrule[1pt] 
	\end{tabular}}
	
\vspace{-0.3cm}
\end{table}

\section{Experiments}
To prove the high performance of PCSA-Net,  we tested it on both synthetic datasets,  SOTS indoor \cite{li2018benchmarking} and Haze4K \cite{liu2021synthetic},  as well as a real dataset,  RTTS \cite{li2019benchmarking}.  The results show that PCSA-Net achieves the best performance in both performance metrics and visual quality.  In the following sections,  we will first introduce the experimental details and datasets,  then analyze the results,  and finally discuss the ablation study. 
\subsection{Implementation Details}
To carry out our experiments,  we utilized the PyTorch 1.11.0 framework,  running on four NVIDIA RTX 4090 GPUs.  Our optimization method of choice was the Adam optimizer,  with the decay rates for $ \beta_{1} $ and $\beta_{2}$ set to 0.9 and 0.999,  respectively.  We started with a learning rate of 0.00015,  which we adjusted using a cosine annealing schedule.  The batch size was configured to be 64.  Drawing from empirical evidence,  we set the penalty parameter $ \lambda$ to 0.2 and $ \gamma $ to 0.25,  training our model for 80,000 iterations.  Additionally,  we employed Contrastive Regularization (CR) \cite{wu2021contrastive} to enhance the quality of dehazed images. 
\subsection{Datasets and Metrics}
We evaluated our proposed PCSA-Net on both synthetic datasets (SOTS indoor
\cite{li2018benchmarking},  Haze4K 
\cite{liu2021synthetic}) and a real-world dataset (RTTS 
\cite{li2019benchmarking}).  Specifically,  we trained on 13,990 image pairs and tested on 500 indoor image pairs from SOTS indoor
\cite{li2018benchmarking}.  Haze4K 
\cite{liu2021synthetic} includes 4,000 image pairs,  with 3,000 used for training and the remaining 1,000 for testing.  Compared to SOTS indoor
\cite{li2018benchmarking},  Haze4K 
\cite{liu2021synthetic} blends indoor and outdoor scenes,  offering a more realistic visual effect.  We also tested the models trained on SOTS indoor
\cite{li2018benchmarking} and Haze4K 
\cite{liu2021synthetic} on the RTTS 
\cite{li2019benchmarking} dataset,  which consists of 1,000 real-world images with dense fog,  demonstrating the robustness and effectiveness of PCSA-Net. 
\subsection{Comparison with State-of-the-art Methods}
\subsubsection{Results on Synthetic Dataset}
Checking out the performance numbers,  Tab.  \ref{performance_SOTS} shows how our PCSA-Net stacks up against the best out there on synthetic datasets.  On the SOTS indoor dataset,  PCSA-Net hits a peak with a PSNR of 39.40dB and an SSIM of 0.991,  which is a PSNR increase of 0.99 over the next top method.  On the Haze4K dataset, PCSA-Net also achieved a PSNR of 33.76 and an SSIM of 0.98, outperforming other methods.  Visually,  as you can see in Fig. \ref{Fig:Haze4K},  
It's clear that the images recovered by PCSA-Net are the closest to the clear images. KDDN \cite{hong2020distilling} and Dehamer \cite{guo2022image} cause the images to be overly enhanced, resulting in dark and discolored images. MSBDN \cite{dong2020multi}  and FFA-Net \cite{qin2020ffa} have a slight haze residue issue, while AOD-Net \cite{li2017aod} leaves larger patches of haze. Surprisingly, our PCSA-Net shows excellent performance in recovering the sky area, which may be due to its ability to perceive long-range relationships effectively. Additionally, because PCSA-Net employs a multi-scale  strategy, it's evident from the images that PCSA-Net recovers the sharpest details and textures.

\subsubsection{Real-world Visual Comparisons}
We tested PCSA-Net on the RTTS \cite{li2019benchmarking} dataset,  which is known for its real-world,  irregular dense fog,  making it a great test for the model's dehazing abilities and robustness.  As shown in Fig. \ref{Fig:RTTS},  the images recovered by our PCSA-Net offer the best visual results in terms of brightness,  texture details,  and color.  Our PCSA-Net removed most of the fog,  resulting in the clearest images.  In contrast,  images restored by other methods like Dehamer \cite{guo2022image},  KDDN \cite{hong2020distilling},  MSBDN \cite{dong2020multi},  FFA-Net \cite{qin2020ffa},  GridDehazeNet \cite{griddehazenet},  and AOD-Net \cite{li2017aod} still had lingering haze,  color distortion,  low brightness,  and lost texture details.  This clearly demonstrates the superior dehazing performance of PCSA-Net. 

\subsection{Ablation Study} 
On the Haze4K dataset,  we conducted an ablation study on various strip attention designs,  including HSA,  VSA,  direct summation of HSA and VSA,  PCSA,  and PCSAM which combines PCSA with a multi-scale  strategy. We progressively enhance the components of our model starting from the `Base'. The term `Base' typically refers to the fundamental U-Net framework equipped with straightforward depth-wise $3 \times 3$ convolutions. 
As shown in Tab. \ref{ablation_netwrok} , the `Base' has a PSNR of 25.76 and an SSIM of 0.92. Introducing HSA can increase the PSNR to 31.76, while introducing VSA can further raise the PSNR to 31.94, demonstrating the effectiveness of both HSA and VSA. If the model simply sums the results after executing HSA and VSA in parallel, the PSNR is 32.96. However, by adopting our designed PCSA, the PSNR can be increased to 33.27. Finally, by combining PCSA with a multi-scale  strategy, we can increase the PSNR by 0.49 to reach 33.76 and achieve an SSIM of 0.98. These results fully validate the effectiveness of our model.
 
\section{Limitation}
While our method effectively captures long-range relationships and maintains low complexity, the large number of parameters may pose challenges for deployment on edge devices. Additionally, the model's operational efficiency may not be real-time, and further improvements are needed.

\section{Conclusion}
In this paper,  we introduce PCSA-Net,  a new image dehazing network that uses a U-Net structure and PCSAB.  PCSAB has PCAS for efficient information gathering in both horizontal and vertical directions.  Each pixel can sense a large area centered around itself,  capturing long-range relationships while keeping the model lightweight.  The multi-scale strategy guides PCAS to dynamically capture contextual information at different scales,  perfectly adapting to various densities and sizes of blur.  We've thoroughly tested PCSA-Net on both synthetic and real datasets,  and the results prove its effectiveness,  efficiency,  and versatility. 
 \section{Acknowledgment}
Xiamen Ocean and Fisheries Development Special Funds (22CZB013HJ04), the Youth Science and Technology Innovation Program of Xiamen Ocean and Fisheries Development Special Funds (23ZHZB039QCB24), the National Natural
Science Foundation of China (Grant No. 62301453).

\bibliographystyle{ACM-Reference-Format}
\bibliography{sample-manuscript}

\appendix

\end{document}